\documentclass[runningheads,a4paper]{llncs}
\usepackage{graphicx}
\usepackage{amsmath,amssymb} % define this before the line numbering.
\usepackage{color}
\usepackage[width=122mm,left=12mm,paperwidth=146mm,height=193mm,top=12mm,paperheight=217mm]{geometry}

\usepackage{times}
\usepackage{soul}
\usepackage{epsfig}
\usepackage{cite}
\usepackage{multirow}
\usepackage{diagbox}
\usepackage{rotating}
\usepackage{overpic}
\usepackage{subcaption}
\usepackage{booktabs}
\usepackage{array}
\usepackage{colortbl}
\usepackage[table]{xcolor}
\usepackage[linesnumbered,ruled]{algorithm2e}

\DeclareMathOperator*{\argmax}{argmax}

\def\ie{\emph{i.e.,~}}
\def\eg{\emph{e.g.,~}}
\def\etal{{\em et al.~}}

\usepackage{array}
\newcolumntype{x}[1]{>{\centering\arraybackslash\hspace{0pt}}p{#1}}

\usepackage{pifont}
\newcommand{\cmark}{\ding{51}}%
\newcommand{\xmark}{\ding{55}}%

\newcommand{\figref}[1]{Fig.~\ref{#1}}
\newcommand{\tabref}[1]{Table~\ref{#1}}
\newcommand{\equref}[1]{Eqn.~(\ref{#1})}
\newcommand{\secref}[1]{Sec.~\ref{#1}}
\newcommand{\myPara}[1]{\vspace{.05in}\noindent\textbf{#1}}
\newcommand{\sArt}{state-of-the-art~}

\graphicspath{{./figs/}}

\newcommand{\addFig}[1]{}
\newcommand{\addFigs}[1]{}
\newcommand*{\affmark}[1][*]{\textsuperscript{#1}}

\usepackage[pagebackref=false,letterpaper=true,colorlinks=true,linkcolor=black,citecolor=black,bookmarks=false]{hyperref}

\begin{document}
% \linenumbers
\pagestyle{headings}
\mainmatter

\title{WebSeg: Learning Semantic Segmentation from Web Searches} % Replace with your title

\titlerunning{WebSeg: Learning Semantic Segmentation from Web Searches}

\authorrunning{Hou \etal}

\author{Qibin Hou\affmark[1] \quad Ming-Ming Cheng\affmark[1] \quad Jiangjiang Liu\affmark[1] \quad Philip H.S. Torr\affmark[2]}
\institute{\affmark[1]Nankai University \quad \affmark[2]University of Oxford \\
\email{andrewhoux@gmail.com, cmm@nankai.edu.cn}}

\maketitle

\begin{abstract}

In this paper, we improve semantic segmentation by
automatically learning from Flickr images associated with a particular keyword,
without relying on any explicit user annotations,
thus substantially alleviating the dependence on accurate annotations when
compared to previous weakly supervised methods.

To solve such a challenging problem, %inspired by human visual mechanism,
we leverage several low-level cues (such as saliency, edges, etc.)
to help generate a proxy ground truth.
Due to the diversity of web-crawled images,
we anticipate a large amount of `label noise' in which
other objects might be present.
We design an online noise filtering scheme
which is able to deal with this label noise, especially in cluttered images.
We use this filtering strategy as an auxiliary module
to help assist the segmentation network in learning cleaner proxy
annotations.
Extensive experiments on the popular PASCAL VOC 2012
semantic segmentation benchmark show surprising good results
in both our WebSeg (mIoU = $57.0\%$)
and weakly supervised (mIoU = $63.3\%$) settings.

\keywords{Semantic segmentation, learning from web, Internet images.}
\end{abstract}

\section{Introduction}

Semantic segmentation, as one of the fundamental computer vision tasks,
has been widely studied.
Significant progress has been achieved
 % as evidenced by rapid performance improvements as demonstrated
on challenging benchmarks,
\eg PASCAL VOC \cite{everingham2015pascal} and Cityscapes \cite{cordts2016cityscapes}.
% and MS COCO \cite{lin2014microsoft}.
%
Existing \sArt semantic segmentation algorithms
\cite{zhao2016pyramid,lin2016refinenet,lin2016efficient}
rely on large-scale pixel-accurate human annotations,
which is very expensive to collect.
To address this problem, recent works focused on
semi-supervised/weakly-supervised semantic segmentation
using user annotations in terms of bounding boxes
\cite{papandreou2015weakly,qi2016augmented},
scribbles \cite{lin2016scribblesup},
points \cite{bearman2016s}, or even keywords \cite{chaudhry2017discovering,hou2016mining,kolesnikov2016seed,wei2016stc,wei2017object,jin2017webly}.
However, using these techniques to learn %how to segment objects in
new categories remains a challenging task %as it still needs to
that requires manually
collecting large sets of annotated data. % albeit less annotations are needed.
Even the the simplest image level keywords annotation might
take a few seconds for a single example \cite{papadopoulos2014training},
which involves substantial human labour, if we consider exploring
millions/billions of images and hundreds of new categories
%in a never ending learning context
in the context of never ending learning \cite{mitchell2015never}.
%
%Taking image-level labels as an example, it
%turns out that around 1 second is needed for labeling
%each class or object category as stated in \cite{papadopoulos2014training}.
%
%Hence, it requires substantial human labors to
%annotate hundreds of thousands of images.

\renewcommand{\addFig}[1]{\includegraphics[width=0.17\linewidth]{teaser/#1}}
\renewcommand{\addFigs}[1]{\addFig{#1.jpg}& \addFig{#1_edge.png} & \addFig{#1_large.png} & \addFig{#1_sal.png} & \addFig{#1_sal_ref.png}}
\newcommand{\addFigsDes}[2]{\begin{sideways} #1 \end{sideways} & \addFigs{#2}}
\begin{figure}[t]
	\centering
	\footnotesize
    \setlength\tabcolsep{1.2pt}
    \begin{tabular}{cccccc} \specialrule{0em}{0pt}{0pt}
        \addFigsDes{dining table}{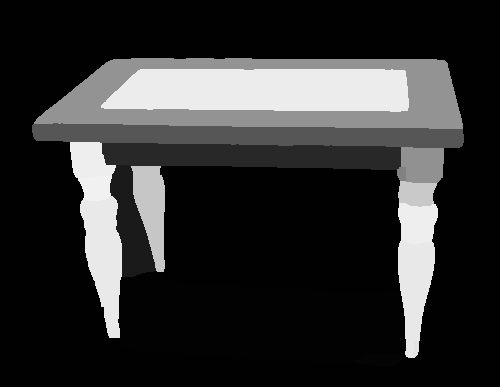}\\
        \addFigsDes{~~~~~bottle}{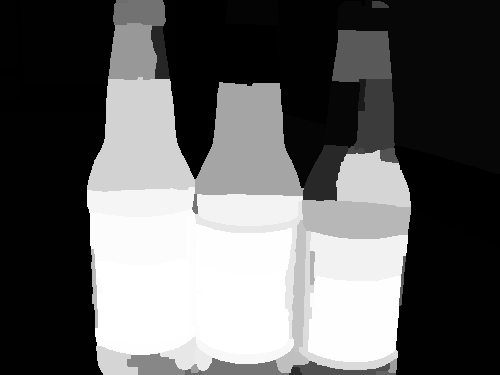} \\
        \addFigsDes{~~~~bicycle}{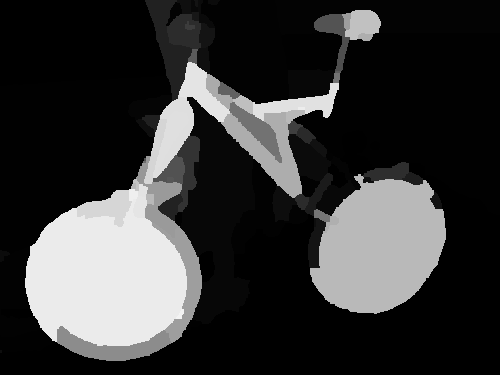} \\
        & (a) images & (b) edges & (c) regions & (d) saliency & (e) proxy GT \\
	\end{tabular}
    %\vspace{-5pt}
	\caption{ (a) Flickr images which are part of the result of searching on a keyword;
        (b) Edge maps produced by the RCF edge detector \cite{liu2016richer};
        (c) Over-segmentation regions derived from (b) using the MCG method \cite{pont2017multiscale};
        (d) Saliency maps generated by the DSS saliency detector \cite{hou2016deeply}; (e) Proxy ground truths that can be directly computed by averaging the
        saliency values for each region. All the low-level cues
        can be used for any category without further training or finetuning.
    }
    \label{fig:teaser}
	%\vspace{-10pt}
\end{figure}

We, as humans, are experts at quickly learning how to identify object regions
% of unknown category
by browsing example images searched using the corresponding keywords.
%
% Our low-level vision ability \cite{wurtz2000central}
Identifying object regions/boundaries and salient object regions
of unknown categories
easily helps us obtain enough pixel-wise `proxy ground truth' annotations.
Motivated by this phenomenon, %considering that existing weakly supervised
%semantic segmentation methods still require more or less human annotations,
% it is straightforward for us to pose such a question:
this paper addresses the following question:
\emph{could a machine vision system automatically learn
semantic segmentation of unseen categories
from web images that are the result of search keyword,
without relying on any explicit user annotations?}
Very recently, some works, such as \cite{hong2017weakly,jin2017webly,wei2016stc},
have proposed various web-based learners attempting to solve the semantic segmentation
problem.
These methods indeed leverage web-crawled images/videos.
However, they assume that each image/frame is associated with correct labels
instead of considering how to solve the problem of label noise presenting 
in the web-crawled images/videos.

\myPara{Our task vs. weakly supervised semantic segmentation.}
Conceptually, our task is \emph{quite different from}
traditional weakly supervised semantic segmentation as we must also deal with label noise.
On one hand, in our task, only a set of target categories are provided
rather than specific images with precise image-level annotations.
This allows our task to be easily extended to cases
where hundreds/thousands of target categories are given
without collecting a large number of images with precise image-level labels.
But this comes at a cost. Due to the typical diversity of web-crawled images,
some of them may contain objects that are
inconsistent with their corresponding query keywords (See \figref{fig:web_images}).
Therefore, how to deal with this \emph{label noise} \cite{frenay2014comprehensive}
becomes particularly important, which is one of our contributions.
%
% This is why our task is more challenging than weakly supervised semantic segmentation.

%In this paper, we are interested in a more challenging problem
%than weakly supervised semantic segmentation,
%\emph{how to learn from unrestricted amount of Internet images to
%build a powerful system that can segment every given categories.}
%
As demonstrated in \figref{fig:teaser},
we show some keyword-retrieved web images containing rich appearance varieties
that can be used as training samples for semantic segmentation.
With the help of %category-agnostic
low-level cues
e.g., saliency, over-segmentation and edges,
we are able to solve the semantic segmentation problem % more generally
without relying on precise image-level noise free labels
required by previous weakly supervised methods
\cite{kolesnikov2016seed,wei2017object,chaudhry2017discovering}, which
required attention cues \cite{zhou2016learning}.
For instance, edge information provides potential locations for object boundaries.
Saliency maps tell us where the regions of interest are located.
With these heuristics, the object regions that correspond to the query keywords
become quite unambiguous for many images, making the unannotated web images a valuable source to
be used as proxy ground truth for training semantic segmentation models.
All the aforementioned low-level cues are \emph{category-agnostic},
making \emph{automatically segmenting unseen categories} possible.
%which means the methods used to generate these cues do not need
%to be trained for the target category, a key property that makes learning for unseen categories possible.
%
%In fact, the whole aforementioned process happens to coincide with the working mechanism of our visual system.

%With the retrieved web images and heuristic cues, another key problem, which is also critical in
%weakly supervised semantic segmentation, is how to deal with the noises
%from the proxy ground truth annotations automatically.
%
Given the heuristic maps produced by imperfect heuristic generators,
to overcome the noisy regions,
previous weakly supervised semantic segmentation
methods \cite{kolesnikov2016seed,wei2017object,krahenbuhl2012efficient,
zheng2015conditional,wei2016stc,hou2016mining}
mostly harness image-level annotations
(e.g. using PASCAL VOC dataset \cite{everingham2015pascal})
to correct wrong predictions for each heuristic map. % based on appearance knowledge from the source images.
In this paper, we consider this ticklish situation from a new perspective,
which attempts to eliminate the negative effect caused by noisy proxy annotations.
To do so, we propose to \emph{filter the irrelevant noisy regions} by introducing an
online noise filtering module.
As a light weight branch,
we embed it into a mature architecture (\eg Deeplab \cite{chen2014semantic}),
to filter those regions with potentially wrong labels.
We show that as long as the most regions are clean,
by learning from the large amount of web data,
we can still obtain good results.

To verify the effectiveness of our approach,
we conduct extensive experiment comparisons on
PASCAL VOC 2012 \cite{everingham2015pascal} dataset.
Given 20 PASCAL VOC object category names,
our WebSeg system %starts to automatically
retrieves 33,000 web images
and immediately learn from these data without any manual assistant.
The experimental results suggested that our WebSeg system,
which does not use manually labeled image,
is able to produce surprisingly good semantic segmentation results (mIOU = $\%57.0$).
Our results are comparable with recently published
state-of-the-art weakly supervised methods,
which use tens of thousands of manually labeled images with precise keywords annotations.
When additional keyword level weak supervision is provided,
our WebSeg system could achieve an mIoU score of $\%63.3$,
which significantly outperforms
the existing weakly supervised semantic segmentation methods.
We also carefully performed a serials of ablation studies to
verify the importance of each component in our approach.

%Compared to weakly supervised semantic segmentation methods,
%when using existing dataset \cite{hou2016mining} containing nearly 24,000 images
%from the ImageNet dataset \cite{russakovsky2015imagenet}
%with well-annotated label-level labels,
%our approach achieves an IoU score of nearly 59\%
%on the popular PASCAL VOC 2012 benchmark \cite{everingham2015pascal}.
%%
%When the training set of PASCAL VOC 2012 \cite{everingham2015pascal} is used,
%we further achieve an IoU score of more than 63\%,
%which is substantially larger than the existing state-of-the-arts.
%%
%When taking in account only 33,000 Internet images,
%we can still obtain great results
%on the PASCAL VOC 2012 benchmark \cite{everingham2015pascal}.
%Experimental results on the popular PASCAL VOC 2012 benchmark \cite{everingham2015pascal}
%are surprising:
%i) our method outperforms \sArt
%weakly supervised methods that are trained with human keywords level annotations;
%ii) and be comparable to fully supervised methods that are
%trained on large-scale pixel-accuracy ground truth annotations!

In summary, our contributions include:
\begin{itemize}
  \item an attractive computer vision task,
    which aims at performing semantic segmentation
    by automatically learning from keyword-retrieved web images.
    % using several low-level cues;
  \item an online noise filtering module (NFM),
  	which is able to effectively filter potentially noisy region labels
    caused by the diversity of web-crawled images.
\end{itemize}

\renewcommand{\addFig}[1]{\includegraphics[height=0.12\linewidth,width=0.16\linewidth]{web_images/#1}}
\begin{figure}[t]
	\centering
	\footnotesize
    \setlength\tabcolsep{1pt}
    \begin{tabular}{cccccc} \specialrule{0em}{1pt}{1pt}
    	\addFig{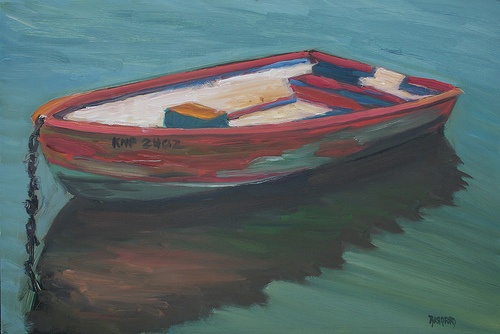} &
        \addFig{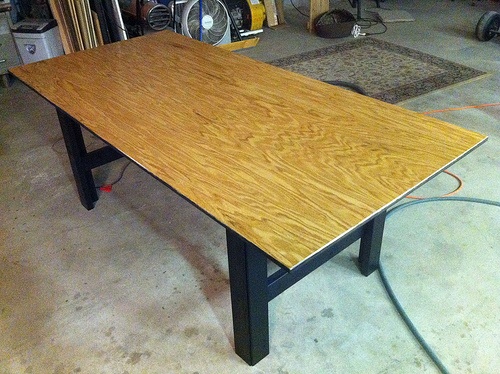} &
        \addFig{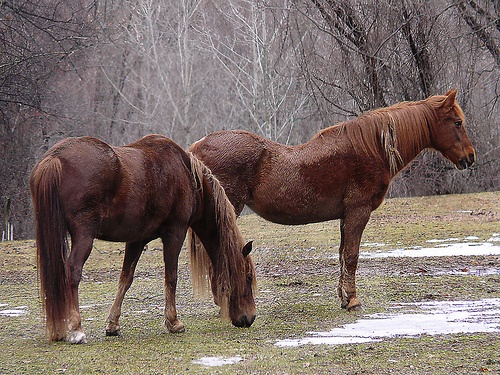} &
        \addFig{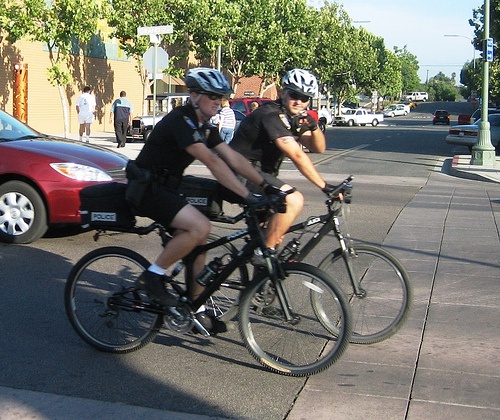} &
        \addFig{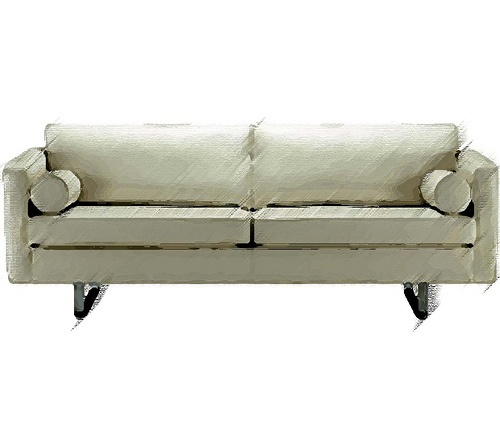} &
        \addFig{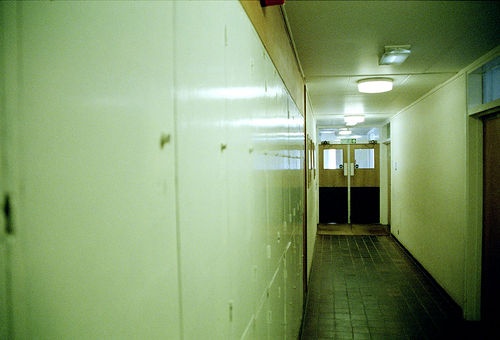} \\  \specialrule{0em}{0pt}{0pt}
        \addFig{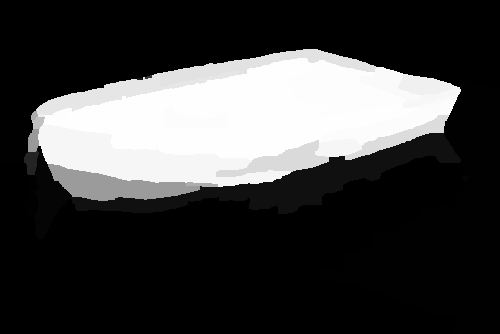} &
		\addFig{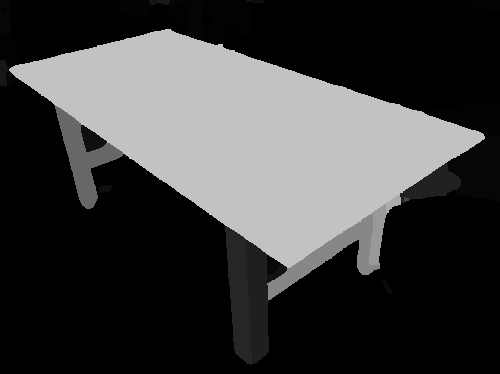} &
        \addFig{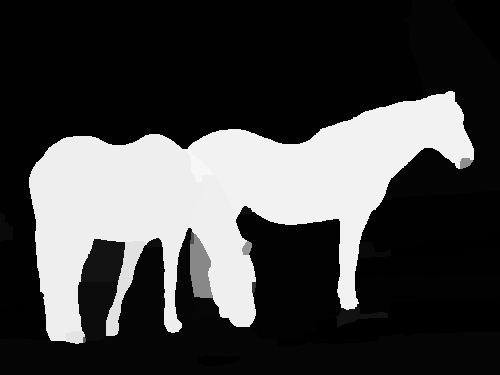} &
        \addFig{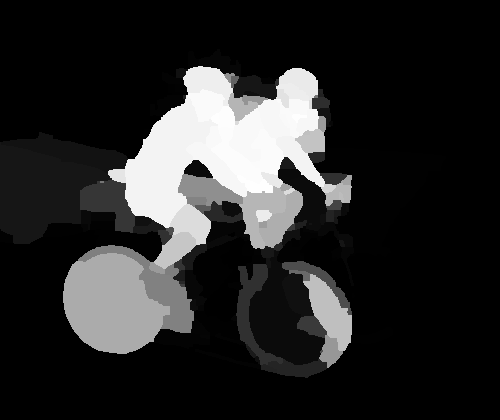} &
        \addFig{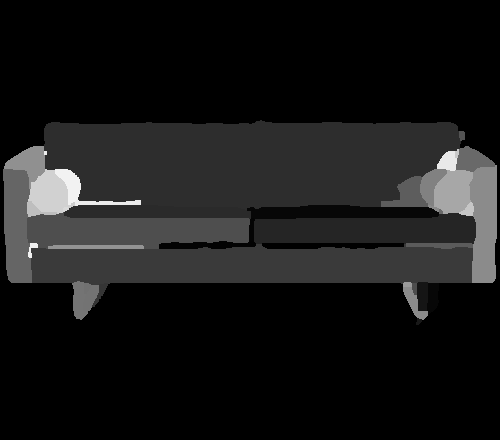} &
        \addFig{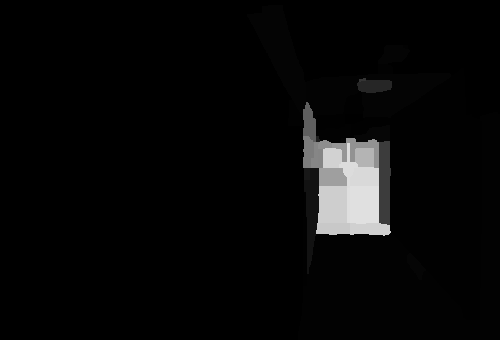} \\  \specialrule{0em}{0pt}{0pt}
        (a) boat & (b) table & (c) horse & (d) bicycle & (e) sofa & (f) potted plant \\
	\end{tabular}
    \vspace{-5pt}
	\caption{A group of images retrieved and the corresponding proxy ground truths. As can be seen, most of the
    semantic objects can be roughly segmented out despite a small number of failing samples.}
	\label{fig:web_images}
    %\vspace{-5pt}
\end{figure}

%\section{Related Works}

%\section{Automatically Learning from the Web-Crawled Images} \label{sec:webly_ss}
\section{Internet Images and Several Low-Level Cues} \label{sec:webly_ss}

The cheapest way to obtain training images is to
download them from the Internet (\eg Flickr).
Given a keyword or more, we are able to easily obtain
a massive number of images that are very likely to be relevant,
without reliable annotations at keyword or pixel level.

\subsection{Crawling Training Images From the Internet} \label{sec:web_images}

Given a collection of keyword, we first download 2,000 images for each category from the Flickr
website according to relevance.
Regarding the diversity of the fetched images, for example, some of them may have
very complex background and some of them may also contain more than
one semantic category (Fig.~\ref{fig:web_images}),
we design a series of filtering strategies to
automatically discard those images with low quality.
To measure the complexity of the web images, we adopt the following schemes.
First, we use the variance of the Laplacian \cite{pech2000diatom,pertuz2013analysis} to determine
whether the input images are blurry.
By convolving an image with the Laplace operator and computing the variance of the resulting map,
we can obtain a score indicating how blurry the image is.
In our experiment, we set a threshold 50 to throw away all the images that are larger than it.
Second, we discard all the images with larger saturation and brightness values by transforming the input images
from RGB color space to HSV space.
This is reasonable as images with lower saturation and brightness are even difficult for humans to recognize.
Specifically, we calculate the mean values of the H and V channels for each image.
If either value is lower than 20, then the corresponding image will be abandoned.
Finally, we get around 33,000 images for training.
%
%We denote these images as $\mathcal{D}(W)$.

\subsection{Low-Level Cues} \label{sec:low_level_cues}

Web-crawled images do not naturally carry object region information
with them.
However, a three-year-old child could easily find precise object regions for most of these images,
even if the corresponding object category has not been explicitly explained to him/her.
%is never been known by him/her.
%
This is mostly because of the capacity of humans processing low-level vision tasks.
Inspired by this fact, we propose to leverage several
low-level cues for the proposed task by mimicking the working mechanism of our vision system.
%
%Previous weakly supervised semantic segmentation methods \cite{hou2016mining,chaudhry2017discovering} mainly
%rely on saliency and attention cues.
%%
%Nonetheless, these cues, in most cases, are not sensitive to the boundaries of real objects, producing
%low-quality predictions.
%
In this paper, we take into account two types of low-level cues, including saliency and edge,
which are also essential in our visual system when dealing with high-level vision tasks.
Salient object detection models provide the locations of
foreground objects as shown in \figref{fig:teaser}(d) but weak knowledge on boundaries.
As a remedy, edge detection models contain rich information about the boundaries, which
can be used to improve the spatial coherence of our proxy ground truths.
Specifically, for saliency detection, we use the DSS salient object detector
\cite{hou2016deeply} to generate saliency maps as done in \cite{hou2016mining}.
For edge detection, we select RCF \cite{liu2016richer} as our edge detector which we found works
better than the HED edge detector \cite{xie2017holistically}.
Furthermore, inspired by \cite{maninis2017convolutional}, we apply the MCG method \cite{pont2017multiscale}
to the edge maps to get high-quality regions.
Some visual results can be found in \figref{fig:teaser}.
As can be observed, with the help of edge information, some undetected salient regions
can be segmented out.
We will show some quantitative results when different heuristic cues are used in \secref{sec:experiments}.
%To make up the weakness of saliency cues on boundaries, here we present to use edge information
%as a constraint to improve the spatial coherence.
%%
%On one hand, like saliency detection, edge detection is actually a class-agnostic task.
%%
%On the other hand, recent edge detectors produce high-quality edge maps, allowing us
%to easily obtain good segmentation maps \cite{pont2017multiscale,arbelaez2011contour}.

Formally, let $I$ denote an image with image labels $\mathbf{y}$, a subset of a predefined set
$\mathcal{L}=\{l_1, l_2, \ldots, l_L\}$, where $L = |\mathcal{L}|$ denotes the number of semantic labels.
We also use $l_0$ to denote the ``background'' category, so we have $\hat{\mathcal{L}}=\{l_0, \mathcal{L}\}$.
Further, let us denote $S$ and $R$ as the corresponding saliency and region maps of $I$, respectively.
%
%We first combine $S$ and $A$ with a pixel-wise maximum operation as suggested in \cite{hou2016deeply},
%yielding an intermediate map $\hat{S}$.
%
Then for each region $R_m \in R$, we perform the following snapping operation with respect to $S$
\begin{equation} \label{eqn:snapping}
H_j = \frac{1}{|R_m|}\sum_{j \in R_m}S_j,
\end{equation}
where $H$ is the resulting heuristic map (\ie the proxy ground truth annotation) used as supervision.
Some visual results can be found in \figref{fig:teaser}(e).

% \subsection{Datasets} \label{sec:datasets}

% %\myPara{Low-Level Cues.}
% %\myPara{Datasets.}
% For convenience, we denote dataset with only simple images (\ie images supposed to have
% one semantic category) as $\mathcal{D}(S) = {\{I_i, \mathbf{y}_i\}}_{i=1}^{N_S}$
% and similarly complex images
% $\mathcal{D}(C) = {\{I_i, \mathbf{y}_i\}}_{i=1}^{N_C}$, respectively,
% where $N_S$ and $N_C$ are the image numbers
% of datasets $\mathcal{D}(S)$ and $\mathcal{D}(C)$.
% %
% Thus, for a complex image $I_i$ from $\mathcal{D}(C)$,
% \cmm{we have $|\mathbf{y}_i| \ge 1$} while when $I_i$ is
% a simple image from $\mathcal{D}(S)$, we have $|\mathbf{y}_i| = 1$.
% %
% In this paper, we classify all the web images as simple images in that our keywords only contain one category
% during each search.

%\myPara{Architecture.}
%We use the notable Deeplab-Large-FOV model \cite{chen2014semantic} as our baseline model as done
%in most previous works \cite{kolesnikov2016seed,wei2016stc,hou2016mining}.
%%
%By ignoring the bottom Conv1-Conv5 layers inherits from the original VGGNet \cite{simonyan2014very},
%we define all the left layers (Conv6, Conv7, Score) as our SSM module.
%%
%Beyond that, we also introduce another branch (RCM) to the end of Conv5 for classifying the category
%of each region.

\section{Online Noise Filtering}

Due to the diversity of the downloaded images,
it is very difficult to extract the desired information from them.
% and ignore noisy regions in the proxy ground truth annotations (See \figref{fig:web_images}).
%%
This section is dedicated to presenting a promising mechanism to solve these problems,
namely online noise filtering module (NFM).

\subsection{Observations}

%It turns out that one of the main issues in \cmm{weakly supervised
%semantic segmentation} is the noisy labels, which are from various weak cues.
Regarding the web-crawled images in our task, like weakly supervised
semantic segmentation, one of the main issues is the noisy labels,
which are from various weak cues.
These low-level cues possibly contain parts of the semantic objects
and even false predictions (\figref{fig:illu_ignore}),
severely influencing the learning process of CNNs and
leading to low-quality predictions.
Furthermore, web-crawled images may contain more than one semantic category.
For example, the images in the `bicycle' category often cover the `person' category
and hence both of them will be predicted to salient regions
with category `bicycle' (\figref{fig:illu_ignore}a).
Previous methods \cite{chaudhry2017discovering,hong2017weakly} solved this problem
using attention models \cite{zhou2016learning}.
%which can \cmm{more precisely} locate the semantic objects given the category.
%
However, the attention models themselves require the supervision of
image-level labels that is impossible to be applied to our task.
%
%Even with image-level supervision,
%it is still hard to eliminate the negative effect brought by noises.
%
Regarding this challenge, a promising way to overcome this is to discard
the noisy regions in the heuristic maps
but meanwhile keep reliable ones unchanged.
To achieve such a goal, in this section,
we present an online noise filtering mechanism
to intelligently filter those noisy regions.

\begin{figure*}[t]
  \centering
  \includegraphics[width=\linewidth]{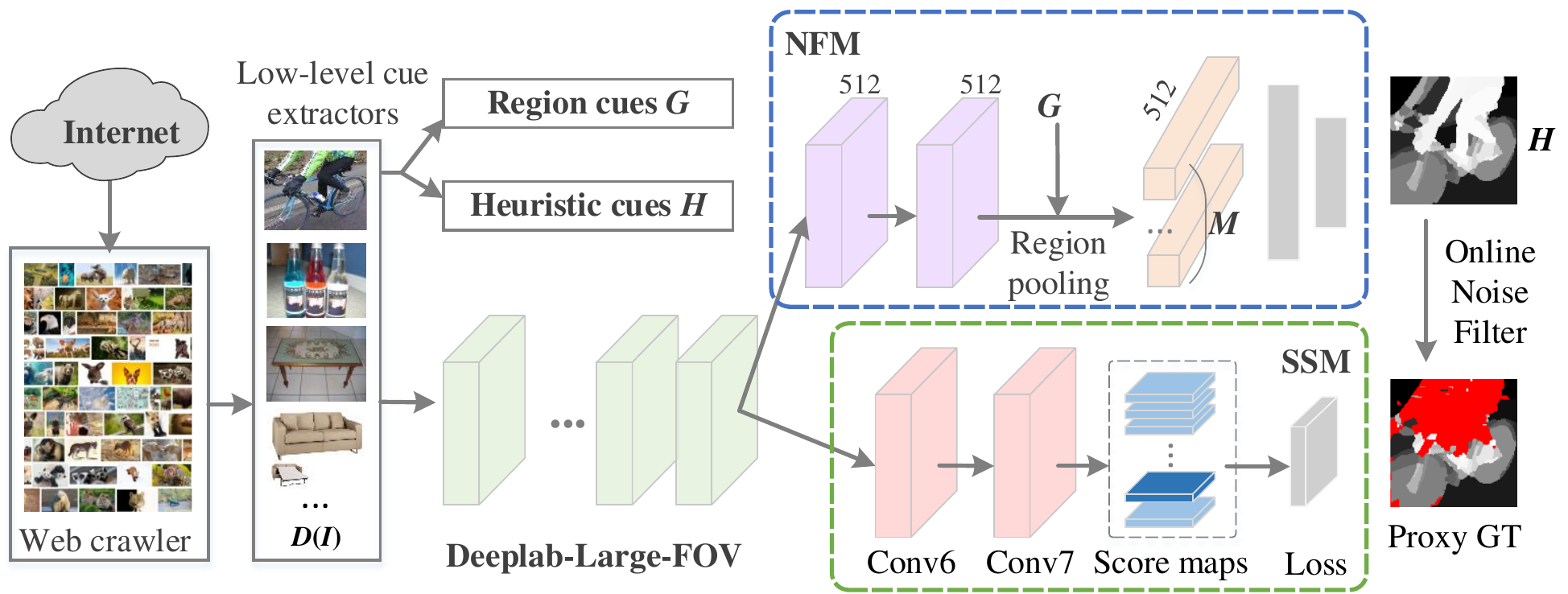}
  \caption{The overall architecture of the proposed method.
    The entire diagram can be separated into three main parts.
    The first one is a web crawler which is responsible for downloading
    Internet images given a collection of user-defined keywords.
    The second part extracts multiple different kinds of
    low-level cues and then combines them
    together as heuristics for learning.
    The last one is a semantic segmentation network with a
    noise filtering module (NFM).
	We use Deeplab-Large-FOV \cite{chen2014semantic}
	as our baseline model in this paper.
  }\label{fig:arch}
\end{figure*}

\renewcommand{\addFig}[1]{\includegraphics[height=0.135\linewidth,width=0.18\linewidth]{ignore/#1}}
\begin{figure}[t]
  \centering
  \small
  \setlength\tabcolsep{1.2pt}
  \begin{tabular}{cccccc}
    \begin{sideways} ~~~~Images \end{sideways} &
    \addFig{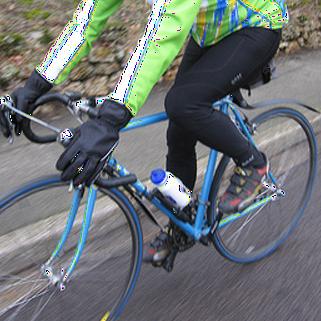} &
    \addFig{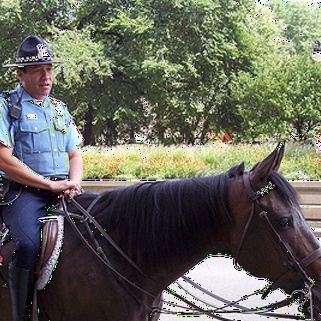} &
    \addFig{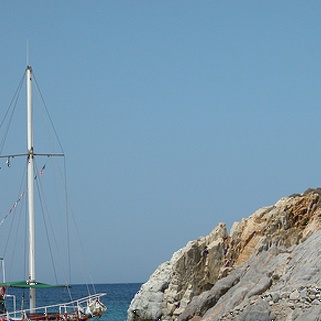} &
    \addFig{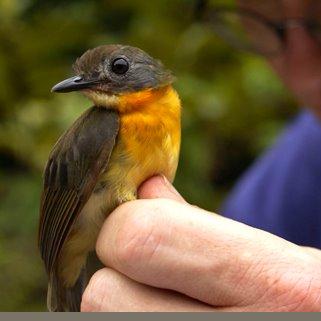} &
    \addFig{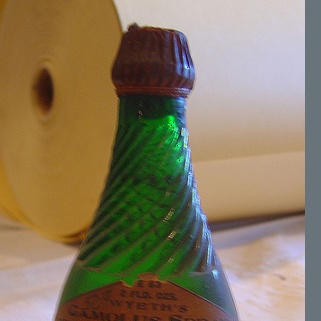} \\
    \begin{sideways} ~~Heuristics \end{sideways} &
	\addFig{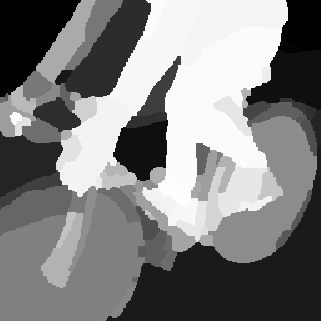} &
    \addFig{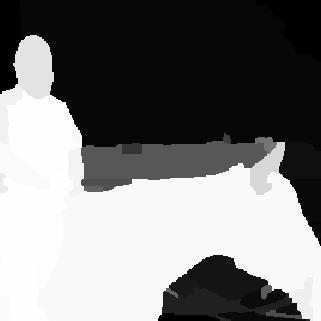} &
    \addFig{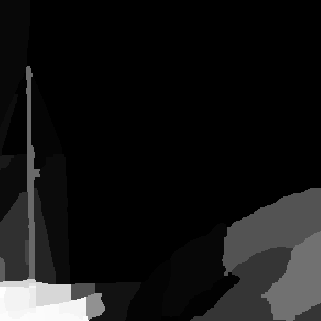} &
    \addFig{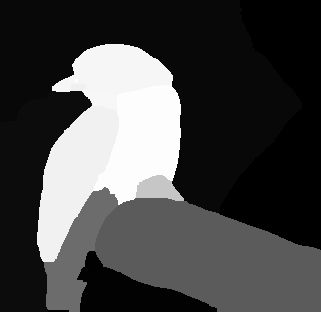} &
    \addFig{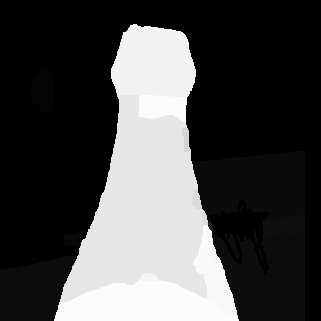} \\
    \begin{sideways} ~~~Results \end{sideways} &
    \addFig{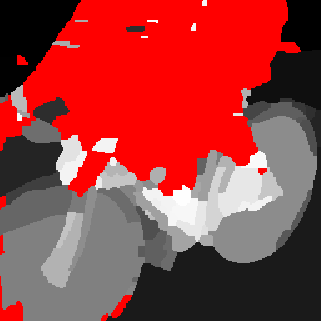} &
    \addFig{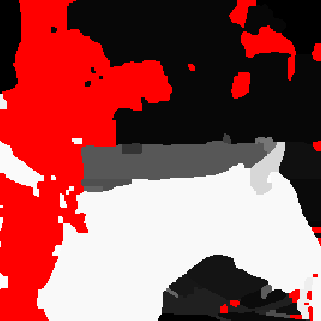} &
    \addFig{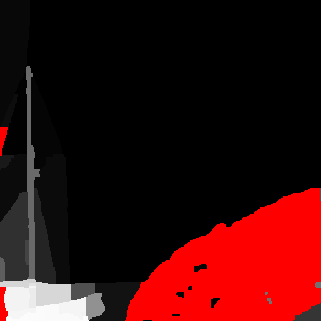} &
    \addFig{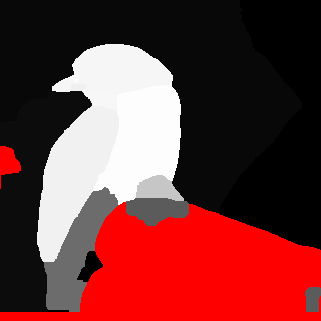} &
    \addFig{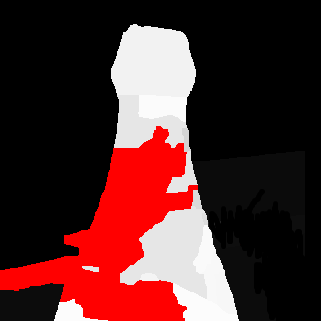} \\
    & (a) bicycle  & (b) horse  & (c) boat & (d) bird & (e) bottle  \\
  \end{tabular}
  \vspace{-5pt}
  \caption{The results extracted from our NFM.
  	Red regions are associated with special labels
    that will be ignored when optimizing SSM.
    (a-b) Ignoring irrelevant categories;
    (c-d) Ignoring undesired stuff from the heuristic maps.
  }\label{fig:illu_ignore}
\end{figure}

\myPara{Overview.}
The pipeline of our proposed approach can be found in \figref{fig:arch}.
%
%The proposed network architecture comprises two main modules,
As the name implies, our proposed scheme is able to filter noisy labels online by introducing
a noise filtering module (NFM) and then uses the generated heuristic maps as the supervision of
the semantic segmentation module (SSM).
%
%Given the category-agnostic low-level cues, we integrate them together to obtain stronger
%heuristic maps that are used as supervision to optimize the CNN model.

%\myPara{Observations.}
%It turns out that one of the main issues in weakly supervised
%semantic segmentation is the noisy labels, which are from various weak (saliency and attention) cues.
%%
%These low-level cues possibly contain parts of the semantic objects
%and even false predictions, severely influencing the learning process of CNNs and hence
%leading to low-quality predictions.
%%
%More importantly, because of the interference of noisy labels, a semantic object might be predicted to
%multiple regions with different categories even though its color is homogeneous.
%%
%For fully supervised semantic segmentation, Chen \textit{et al.} and Zhao \textit{et al.} presented versatile
%pyramid structures \cite{chen2017deeplab,zhao2016pyramid} to increase the receptive fields of CNNs so as to
%address the problem of non homogeneous regions.
%%
%However, the improvement by such a solution in the case of weakly supervised semantic segmentation seems
%trivial owing to the inaccurate ``ground truth''.
%%
%Fortunately, edges provide us more reliable information, which can also be easily transformed to regions.
%%
%A possible way to improve the results from CNNs might be snapping them with reliable
%regions \cite{maninis2017convolutional}, but the improvement is not obvious as well.

\subsection{Online Noise Filtering}

\myPara{Noise Filtering Module (NFM).}
Given an image $I$ and its region map $R$,
it is easy for us to predict the label of each region $R_m$.
Specifically, the first part of our NFM consists of two
convolutional layers for building higher-level
feature representations, both of which have 512 channels, kernel size 3, and stride 1.
Then, we introduce a channel-wise region pooling layer following
the formulation of \equref{eqn:snapping}
to extract equal number of features from each region.
%
%But differently, for each region $G_i$ in a feature map, we only output the average value inside it.
%
Suppose there were totally $M$ regions in image $I$, the dimensions of the output of the
channel-wise region pooling layer would be $(M, 512, 1, 1)$, \ie the mini-batch is $M$.
Finally, two fully connected layers are added as the classifier, which are with
1,024 and $|\mathcal{L}|$ neurons, respectively.
Notice that all the weighted layers in our NFM are followed by ReLU layers for non-linear
transformation apart from the last one.
Please note that designing more complex structure might be more powerful here
but this is beyond the scope of this paper as our goal is to show how to learn to discard those regions
with potentially noisy labels.

\myPara{Learning to Filter Noisy Labels.}
%Let us first consider a simple case where the inputs are all simple images.
%
Given an image $I$ with image-level label $\mathbf{y}$ and its corresponding region map $R$
and heuristic map $H$, for any region $R_m$, if $\sum_{j \in R_m} {H_j} > 0$ then
we say that region $R_m$ belongs to foreground (otherwise background).
During training, the ground truth label of each foreground region should be $\mathbf{y}$.
%Each foreground region should be one of the category in $\mathbf{y}$ and its ground truth label .
%
%For background regions, the ground truth labels are set to $l_0$.
%
%In SSM,
For background regions, we give them a special label
$l_s (l_s \notin \hat{\mathcal{L}})$, indicating that
these regions should be ignored during training as we only care about foreground in NFM.
%
%For each foreground region $G_i$, let $p_i^{(l)} = P_i(l \in \mathcal{L})$ be the probability of $G_i$ predicted to be label $l$ and let $L_i = \argmax{\mathbf{p}_i}$
%For each foreground region $G_i$,
Let us denote $f^m(I)$ as the activations corresponding to $R_m$ in the score layer of NFM.
% and $p^i_l$ the probability of $G_i$ predicted to be label $l$.
%
Notice that we omit the network parameters here for notational convenience.
The predicted label of $R_m$ can be obtained by
$C_m = \argmax_{l \in \mathcal{L}}{f^m_l(I)}$.
If $C_m \notin \mathbf{y}$, then $H_{j, j \in R_m}  = l_s $.
%the corresponding pixels in $H$ will be annotated as $l_s$,
%\ie back-propagating zero gradients for each pixel in $G_i$ in SSM.
%
The resulting $H$ will be used as the annotation of SSM.
The bottom row in \figref{fig:illu_ignore} shows some typical examples of $H$.
Compared to the middle row in \figref{fig:illu_ignore}, we can observe that
our online noise filtering module indeed helps a lot when dealing with images
with interferential categories.
%
%Although some desired regions are also filtered sometimes, this case hardlyoccurs

\myPara{Semantic Segmentation Module (SSM).}
%The structure of our SSM is the same to Deeplab-Large-FOV \cite{chen2014semantic} except the
%loss function.
The proxy annotations sent to SSM are gray-level continuous maps, each pixel of which indicates
the probability of being foreground.
Therefore, following \cite{wei2016stc}, we use the following cross-entropy loss
to optimize our SSM
\begin{equation} \label{eqn:ce_loss}
\mathcal{E}(\theta) = \sum_{n=1}^N
\left( \hat{p}_n^0 \log p_n(l_0|I; \theta) + \hat{p}_n^c \log p_n(\mathbf{y}|I; \theta)
\right),
\end{equation}
where $N$ is the number of elements in $I$, $\theta$ is the network
parameters, $\hat{p}_n^c = H_n$, denoting the probability of the $n$th
element being salient, $\hat{p}_n^0 = 1 - \hat{p}_n^c$, and $p_n(\mathbf{y}|I; \theta)$ the probability of
the $n$th element belonging to $\mathbf{y}$, which is from the network prediction scores.

\myPara{Training.} During the training phase, we found that the NFM module always tends to over-fit
when it reaches convergence.
The classifier will predict the same label to almost all the foreground regions influenced by the noisy labels,
which weakens the role of our NFM.
To address this problem, we change the learning rate of the weighted layers in NFM by multiplying
a fixed factor 0.1 to convolutional layers and 0.01 to fully connected layers.
We empirically found that such an under-fitting state of NFM
makes our whole model work the best.
%
%Notice that we do not back-propagate the gradients from RCM to the main
%trunk of CNN as we found this may have an negative effect on the performance of the SSM module,
%resulting in a nearly 1 percent decrease in terms of the mean IoU score.
%
During the inference phase,
we discard NFM while only keep the original layers in Deeplab.
Therefore, we do not introduce any additional computation into the
testing phase of Deeplab model.

\subsection{Refinement}

\begin{algorithm}[tb]
\caption{Generating ``ground truth'' for Refinement Stage}
\label{alg:update_gt}
    \SetKwInOut{Input}{Input}
    \SetKwInOut{Output}{Output}
    \Input{Image $I$ from the training set; Image labels $\mathbf{y}$;
        Prediction scores $g(I)$}
    \Output{$T$}
    %$T = zeros (N)$, $N$ is the number of pixels\;
    \For {\text{each pixel} $j \in I$} {
        %$T(j) = \argmax_{l \in \mathbf{y}}{g_l(j)}$\;
        $\hat{g}(j) \leftarrow \sigma(g(j))$, $\sigma$ is the softmax function \;
        $\hat{T}(j) \leftarrow \frac{1}{Z(j)} \hat{g}(j) \cdot \mathbf{y}$, $Z(j)$ is the partition function\;
        %$M(m) = h ({\bf s}(m), {\bf a}(m))$ \;%\COMMENT{$M(m) \in [0,1]$}
    }
    $\tilde{T} \leftarrow \text{CRF}(\hat{T})$ \;
    $T \leftarrow \argmax_{l \in \mathbf{y}}{\tilde{T}_l}$ \;

\end{algorithm}

With the above CNN trained based on heuristic maps,
it is possible for us to iteratively train more powerful CNNs by
leveraging the supervision from the keywords as done in \cite{wei2016stc,hou2016mining}.
Suppose in the $r$-th iteration of the learning process the network parameter
is denoted by $\theta^r$.
We can use $\theta^r$ to generate the prediction scores $g^r(I)$ of SSM.
Given an image $I$, let $\mathbf{y}$ be its image-level labels.
The segmentation map $T^r$ of $I$ can be computed according to Alg.~\ref{alg:update_gt}.
With $T^r$, we are able to optimize another CNN (the Deeplab model here),
which may give us better results.
Notice that after the first-round iteration, we do not use the NFM here any more
as $T^r$ already provides us more reliable `ground truth.'
By carrying out the above procedures iteratively, we can gradually refine the
segmentation results.

\section{Experiments} \label{sec:experiments}

%\Todo{
%To do
%\begin{itemize}
%  \item Using the new CRF model.
%  \item Comparison with web-based methods.
%  \item Showing more results on new categories.
%\end{itemize}
%}

Our autonomous web learning of the semantic segmentation task is similar
to previous weakly supervised semantic segmentation task but now
we can deal with label noise.
To better verify the effectiveness of our proposed method, in this section, we compare our proposed approach
with existing weakly supervised semantic segmentation methods and meanwhile analyze the
importance of each component in our approach by ablation experiments.
Differently from our new task which leverages only category-independent cues,
traditional weakly supervised semantic segmentation methods also considers the knowledge of precise image-level labels
(e.g., attention cues).
For fair comparisons, we separate our experiments into two groups, which respectively represent
weakly supervised semantic segmentation (\emph{including attention cues}) and semantic segmentation
\emph{with only category-independent cues}.

\subsection{Implementation Details}

\myPara{Datasets.}
In our experiment, we only use one keyword for each search.
We denote the collected dataset with only web images as $\mathcal{D}(W)$,
thus each image in $\mathcal{D}(W)$ is supposed to have only one semantic category.
Furthermore, to compare our approach with existing weakly supervised methods,
we choose the same datasets as in \cite{hou2016mining}, which contain two parts.
The first dataset is similar to our $\mathcal{D}(W)$, containing images
with only single image-level labels, which are originally
from the ImageNet dataset \cite{russakovsky2015imagenet}.
Here, we denote it as $\mathcal{D}(S)$.
The second part, denoted as $\mathcal{D}(C)$, has only images from the PASCAL VOC 2012
benchmark \cite{everingham2015pascal} plus its augmented set \cite{hariharan2011semantic}.
More details can be found in \tabref{tab:datasets}.
We evaluate our approach on the PASCAL VOC 2012  benchmark and report the results
on both the 'val' and 'test' sets.

\myPara{Low-Level Cues.} Other than saliency cues and region cues, it is reasonable to
harness attention cues \cite{zhang2016top} for weakly supervised semantic segmentation.
Let us denote $A$ as the attention cues of image $I$.
We first perform a pixel-wise maximum operation between $A$ and $S$ (saliency cues)
as in \cite{hou2016mining}, which
aims to preserve as many heuristic cues as possible.
Then, we use Eqn.~\ref{eqn:snapping} to perform a snapping operation
with respect to the combined map, yielding
the heuristic cues for training our first-round CNN.
%
% For datasets, we use the simple dataset provided by \cite{hou2016mining}, which includes around
% 24,000 images from ImageNet \cite{russakovsky2015imagenet} ($\mathcal{D}(S)$) plus an augmented
% PASCAL VOC 2012 dataset \cite{everingham2015pascal,hariharan2011semantic} ($\mathcal{D}(C)$)
% for weakly supervised semantic segmentation experiments and pure web images ($\mathcal{D}(W)$)
% for our new task (See Table~\ref{tab:datasets}).
%

\myPara{Model Settings and Hyper-Parameters.}
We use the publicly available Caffe toolbox \cite{jia2014caffe} as our implementation tool.
Like most previous weakly supervised semantic segmentation works,
we use VGGNet \cite{simonyan2014very} as our pre-trained model.
The hyper-parameters we used in this paper are as follows: initial learning rate (1e-3),
divided by a factor of 10 after 10 epochs, weight decay (5e-4), momentum (0.9), and mini-batch
size (16 for the first-round CNN and 10 for the second-round CNN).
We train each model for 15 epochs.
We also use the conditional random field (CRF) model proposed in \cite{lin2016scribblesup}
as a post-processing tool to enhance the spacial coherence, in which the graphical model
is built upon regions.
This is because we already have high-quality region cues which are derived from edge knowledge.
Instead of both color and texture histograms as in \cite{lin2016scribblesup},
we only use the color histograms for measuring the similarities between adjacent regions
as our regions are derived from edge maps which already contain semantic information.
All the other settings are the same to \cite{lin2016scribblesup}.

\newcommand{\xiaowuhao}{\fontsize{8pt}{\baselineskip}\selectfont}

\begin{table}[t!]
    \centering
    \xiaowuhao
    \renewcommand{\arraystretch}{.8}
    \caption{Dataset details and default dataset settings in each iteration.
    	As WebSeg relies on only category-agnostic cues,
    	the training images from PASCAL VOC dataset cannot be used.
        Because of space limitation, we use `weak' to represent weakly supervised
        setting.
    }\vspace{5pt}
    \renewcommand{\tabcolsep}{1mm}
    \begin{tabular}{cccccc} \toprule[1pt]%\whline{1pt}%\cline{1-2}
	   Datasets & \#Images & single label & precise label & source & task  \\ \midrule[1pt]
        $\mathcal{D}(S)$ & 24,000 & \cmark & \cmark & ImageNet \cite{russakovsky2015imagenet} & weak (rounds 1 \& 2 ) \\ \midrule[-2pt]
        $\mathcal{D}(C)$ & 10,582 & \xmark & \cmark & PASCAL VOC 2012 \cite{everingham2015pascal} & weak (round 2)\\ \midrule[-2pt]
        $\mathcal{D}(W)$ & 33,000 & \cmark & \xmark & pure web & WebSeg (rounds 1 \& 2) \\
        %CCN & - & 38.2\% & - & 39.6\% \\  \midrule[-0.7mm]
        \bottomrule[1pt]
        \end{tabular}
        \label{tab:datasets}
        %\vspace{-10pt}
\end{table}

\newcommand{\thickhline}{%
    \noalign {\ifnum 0=`}\fi \hrule height 1pt
    \futurelet \reserved@a \@xhline
}

\newlength\savedwidth
\newcommand{\whline}[1]{\noalign{\global\savedwidth\arrayrulewidth \global\arrayrulewidth #1}%
                   \hline \noalign{\global\arrayrulewidth\savedwidth}}

\newcolumntype{"}{@{\hskip\tabcolsep\vrule width 0.6pt\hskip\tabcolsep}}

\begin{table*}[tp]
  \centering
  \xiaowuhao
  \caption{Ablations for our proposed approach.
  	Notice that all the results are directly from our CNNs without
  	using any post-processing tools unless noticed.
  	The best result in each case are highlighted in \textbf{bold}.
  	Inferior numerals represent the growth to the base numbers.
  	All the results are measured on the PASCAL VOC 2012 validation set.
  }
  \vspace{5pt}
  \begin{subtable}[t]{2.15in}
    \begin{tabular}{x{19mm}|x{16mm}x{16mm}} %\toprule[1pt]
	  & \multicolumn{2}{c}{First round IoU (\%)} \\ \whline{1pt}%\midrule[0.75pt]%\cline{1-2}
      Low-Level Cues & Weak & WebSeg \\ \whline{1pt}
	  sal &  $52.59$ & $49.92$\\
	  %sal + att &  $54.49_{+1.90}$ \\
      \textbf{sal + edge} & $54.39_{+1.80}$ & $\mathbf{52.09_{+2.17}}$ \\
      \textbf{sal + att + edge} & $\mathbf{56.08_{+3.49}}$ & --- \\
    \end{tabular}
    \subcaption[]{\textbf{Low-level cues:}
      Results when different low-level cues are used during the first-round iteration.
      For space convenience, we use the abbreviations of the cue names if needed.
      As can be seen, more low-level cues do help in both tasks.
    }\label{table:low_level}
    \vspace{2mm}
  \end{subtable}
  \quad
  \begin{subtable}[t]{2.5in}
    \begin{tabular}{x{10mm}|x{11mm}|x{18mm}|x{20mm}} %\whline{1pt}%\cline{1-3}
	  NFM? & Region & Weak IoU (\%) & WebSeg IoU (\%) \\ \whline{1pt}%\cline{1-3}
	  \xmark & - &  $56.08$ & 52.09\\
	  \cmark & Large & $56.46_{+0.38}$ & - \\
      \cmark & Medium & $56.54_{+0.46}$ & - \\
      \cmark &  \textbf{Small} & $\mathbf{57.00_{+0.92}}$ & $\mathbf{53.89_{+1.80}}$ \\
    \end{tabular}
    \subcaption[]{\textbf{The role of our NFM:}
      We use ``Small'', ``Medium'', and ``Large'' to represent the different sizes
      of regions derived from edges \cite{liu2016richer} by MCG \cite{pont2017multiscale}.
      Our NFM is especially useful when dealing with more noisy proxy ground truths
      (the right column).
    }\label{table:rcm_module}
  \end{subtable}
\end{table*}

\subsection{Sensitivity Analysis}

To analyze the importance of each component of our approach, we perform a number of
ablation experiments in this subsection.

\myPara{Low-Level Cues.}
The proposed learning paradigm starts with learning simple images and then
transitions to general scenes following \cite{wei2016stc}.
In this paragraph, we analyze the roles of different low-level cues
and their combinations in the first-round iteration.
For fair comparisons, we discard the NFM here whereas only keep the SSM.
\tabref{table:low_level} shows the quantitative results of
using different low-level cues.
With only saliency cues,
our first-round CNN achieves a baseline result with a mean IoU score of 52.59\%.
With the edge information incorporated, the baseline result can be improved
by 1.8\% in terms of mean IoU.
When all three kinds of low-level cues are considered, we achieve the best results,
which is 56.08\%, around 3.5\% improvement compared to the baseline result.
Similar phenomenon can also be found when only utilizing web images in Table~\ref{table:low_level}.
These results verify the effectiveness of the low-level cues we leverage.
By comparing the results on the two tasks in Table~\ref{table:low_level},
we can observe that under the same experiment setting the web-crawled images
are more complex than the images from the ImageNet.
This is reasonable because unlike the ImageNet images the web-crawled images
are not with accurate labels (See \figref{fig:web_images}).

\myPara{The Role of NFM.}
Table~\ref{table:rcm_module} lists the results when the NFM is used or not during the phase of training the first-round CNN.
One can observe that with small regions provided, using NFM gives a 0.92\% improvement
compared to not using it and a 0.38\% improvement compared to using large regions.
Actually, the most important difference between whether using NFM lies in the ability of predicting images
with multiple categories.
As shown in \figref{fig:visualComp} (columns c and d), our NFM is able to erase those noisy
regions caused by the wrong predictions of saliency model and hence produces cleaner results.
This phenomenon is especially obvious when training on web images.
As shown in Table~\ref{table:rcm_module}, there is an improvement of 1.8\% on the 'val' set
during the first-round iteration compared to the baseline result without NFM.
Compared to the growth range (+0.92\%) with weak supervision, we can observe the noise-erasing
ability of our NFM is essential, especially when handling noisy web images.
From a visual standpoint, \figref{fig:illu_ignore} provides more intuitive and direct results.
Comparing the bottom two lines of \figref{fig:illu_ignore}, we can observe that
our method after NFM successfully filters most noisy regions (undesired categories and
stuffs in the background) in the heuristic cues.
This makes our segmentation network learn cleaner knowledge from the heuristic cues.

%Furthermore, the cleaner segmentation maps produced by SCN with RCM allows our CCN to generate far
%better results, yielding 1.25 points performance gain, which is 0.33 points higher than the first round
%(Table~\ref{table:rcm_module}).

\myPara{Region Size in NFM.}
The region sizes play an important role in our NFM.
Here, we compare the results when using different region sizes in our NFM.
The regions are directly derived from the edge maps \cite{liu2016richer}.
By building the Ultrametric Contour Maps (UCMs) \cite{pont2017multiscale} with different thresholds,
we are allowed to obtain regions with different scales.
Here, we show the results when setting threshold to 0, 0.25, and 0.75, which correspond to small, medium, and large
regions, respectively.
The quantitative results can be found in Table~\ref{table:rcm_module}.
As the number of regions increases, our approach achieves higher and higher mean IoU scores
on the PASCAL VOC 2012 validation set.
Training with small regions enables $0.54$ percent improvement compared to using large regions.
The reason for this might be that lower thresholds provide more accurate regions, making less regions
stretch over more than one category.

\renewcommand{\xiaowuhao}{\fontsize{8pt}{\baselineskip}\selectfont}

\begin{table}[tp]
    \centering
    \xiaowuhao
    \caption{\textbf{Results based on different training set:} As can be seen, our approach
    trained on only simple images has already achieved an IoU score of 58.79\% for weakly
    supervised semantic segmentation. With the VOC training set incorporated, an IoU score
    of 60.45\% can be obtained without using any post-processing tools, which is already better
    than most of the existing methods (See \tabref{tab:comps}). When using images crawled from the Internet, we can also
    achieve a score of 54.78\%.}
    \label{table:training_set}
    \vspace{5pt}
    \begin{tabular}{x{28mm}|x{20mm}|x{30mm}|x{20mm}} %\whline{1pt}
        \multicolumn{2}{c|}{Weakly} & \multicolumn{2}{c}{WebSeg}\\ \whline{1pt}
	        Train Set \& \#Images & IoU (\%) & Train Set \& \#Images & IoU (\%) \\ \whline{1pt}
	        $\mathcal{D}(S)$, ~24,000 & $58.79$ & $\mathcal{D}(W)$, 33,000 & 54.78 \\
	        $\mathcal{D}(C)$, ~10,582 & $59.03_{+0.24}$ & $\mathcal{D}(C)$, 10,582 & --- \\
            \textbf{$\mathcal{D}(S+C)$,~34,582} & $\mathbf{60.45_{+1.66}}$ & $\mathcal{D}(W+C)$, \textbf{43,582} & $\mathbf{57.39_{+2.61}}$ \\ %\whline{1pt}
    \end{tabular}
    \vspace{-10pt}
\end{table}

\begin{table}[tp]
    \centering
    \xiaowuhao
    \caption{\textbf{Results w/ and w/o CRF model:} It is obvious that
        the CRF model greatly improves the results on both weakly supervised setting and WebSeg.}
    \label{table:crf_results}
    \vspace{5pt}
    \begin{tabular}{x{22mm}|x{19mm}x{19mm}|x{19mm}x{19mm}} %\toprule[1pt]%\whline{1pt}%\cline{1-4}
		~ & \multicolumn{2}{c|}{First Round IoU (\%)} & \multicolumn{2}{c}{Second Round IoU (\%)} \\ \whline{1pt}
        CRF Method & Weakly & WebSeg & Weakly & WebSeg \\ \whline{1pt}
		\xmark &  $57.00$ & $53.89$ & 60.45 & 54.78\\
		%CRF & $59.29_{+2.29}$ & $55.17_{+1.28}$ & $62.16_{+1.71}$ &
        %$56.44_{+1.66}$\\
        CRF & $\mathbf{60.56_{+3.56}}$ & $\mathbf{56.58_{+2.69}}$ & $\mathbf{63.14_{+2.69}}$ & $\mathbf{57.10_{+2.32}}$ \\ %\bottomrule[1pt]%\whline{1pt}%\cline{1-4}
        %2 & Region-Based & $\mathbf{61.39}$ \\ \whline{1pt}%\cline{1-4}
	\end{tabular}
    \vspace{-5pt}
\end{table}

\myPara{The Role of CRF models.}
Conditional random filed models have been widely used in semantic segmentation tasks.
In Table~\ref{table:crf_results}, we show the results when the CRF is used or not.
Obviously, the CRF model does help in all cases.

\myPara{The Role of Different Training Sets.}
We try to use different training sets to train the second-round CNN
while keep all the setting of the first-round one unchanged.
As can be seen in Table~\ref{table:training_set}, combining both $\mathcal{D}(S)$ and $\mathcal{D}(C)$
allows us to obtain the best results, which is 1.66 points higher than only using $\mathcal{D}(S)$ for
training.
For pure web supervision, incorporating the images of $\mathcal{D}(C)$ is also helpful, which leads to
a 2.61-point extra performance gain.
This indicates that more high-quality training data does help and complex images provide more information
between different categories that would be useful for both weakly supervised semantic segmentation and semantic
segmentation with only web supervision.

\renewcommand{\addFig}[1]{\includegraphics[width=0.16\linewidth]{vis_comp/#1}}
\renewcommand{\addFigs}[1]{\addFig{#1.jpg} & \addFig{#1.png} & \addFig{#1_seg.png} &
			\addFig{#1_scn.png} & \addFig{#1_scn_crf.png}
            & \addFig{#1_ccn_crf.png} }
\begin{figure}[t]
	\centering
    \scriptsize
	\setlength\tabcolsep{0.4mm}
	\begin{tabular*}{\linewidth}{cccccc}
       \addFigs{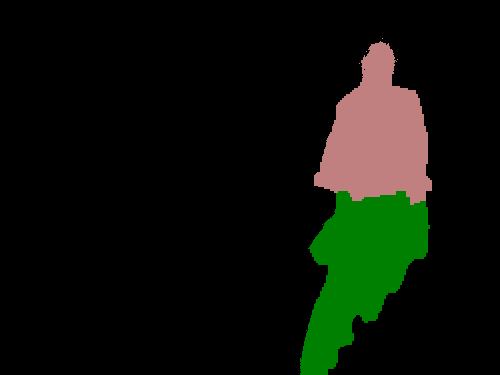}\\
       \addFigs{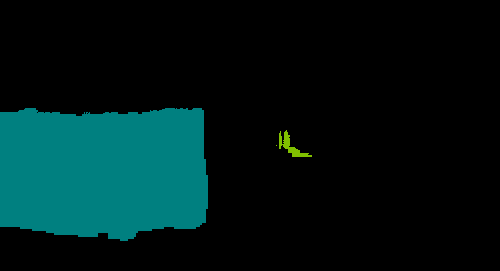}\\
       \addFigs{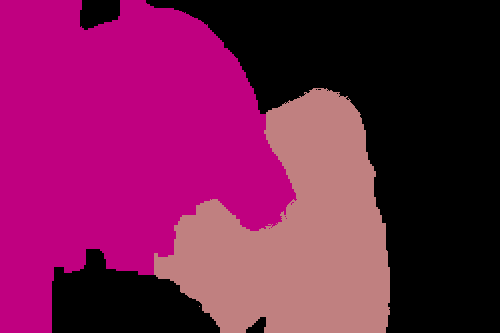}\\ \midrule[0mm]
       \addFigs{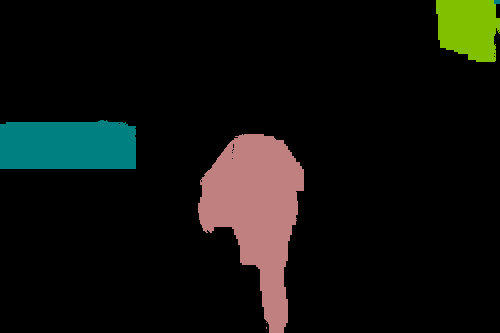}\\
       \addFigs{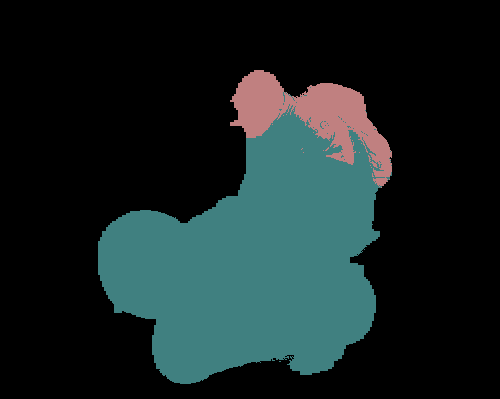}\\
       \addFigs{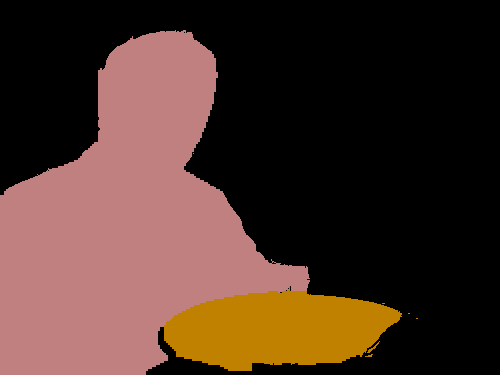}\\
       Source & GT & 1st round w/o NFM & 1st round w/ NFM & 1st round + CRF & 2nd round + CRF \\
	   (a) & (b) & (c) & (d) & (e) & (f) \\
	\end{tabular*}
	\vspace{-5pt}
	\caption{Visual comparisons for weakly-supervised semantic segmentation using different settings of our proposed method. All the images come from the validation set of PASCAL VOC 2012 segmentation benchmark. Other than showing only successful results (the top 3 rows), we also show failure cases of our method (the bottom 3 rows).}
	\label{fig:visualComp}
	%\vspace{-10pt}
\end{figure}

\subsection{Comparisons with Existing Works}

In this subsection, we compare our approach with the existing state-of-the-art weakly supervised semantic segmentation works.
All the methods compared here are based on the Deeplab-Large-FOV baseline model \cite{chen2014semantic}
except special declarations.

Table~\ref{tab:comps} lists the results of prior methods and ours on both `val' and `test' sets.
As can be found, with weak supervision, our approach achieves a mean IoU score of $60.5\%$ on the `val'
set without using CRF.
This value is already far better than the results of all prior works no matter whether the CRFs are used for them.
By using CRFs as post-processing tools, our method realizes the best results on the `val' and `test' sets,
which are both higher than 63.0\%.
%Compared to DCSP-ResNet \cite{chaudhry2017discovering}, our method obtains a mean IoU score of 63.1\% when using region-aware CRF, which is 2.3\% performance gain higher than it and more than 4\% performance gain higher than all the other works.
%
%Similar phenomenon can also be observed when testing on the `test' set.
When considering training on web images with only category-agnostic cues, our results on
both the `val' and `test' sets are also better than most previous works and comparable
to the state-of-the-arts.
%
%Compared to a similar work STC to our task \cite{wei2016stc},
This reflects that despite extremely noisy web images, our approach is robust as well.
This is mostly due to the effectiveness of our proposed architecture, the ability of
eliminating the effect of noisy heuristic cues.

%\vspace{-10pt}
\begin{table}[t!]
    \centering
    %\scriptsize
    %\footnotesize
    \xiaowuhao
    \caption{Quantitative comparisons with the existing
            \sArt approaches on both `val' and `test' sets.
            The best result in each column is highlighted in \textbf{Bold}.
            For fair comparisons, we split the table into two parts according
            to the type of supervision. We also abbreviation `weak' to denote
            weakly supervised approaches and `web' the approach with web
            supervision. `$\dagger$' stands for our results with weak supervision.}
    \vspace{5pt}
    \begin{tabular}{x{43mm}x{10mm}x{10mm}x{15mm}x{15mm}x{20mm}} \hline %\toprule[0.5pt]%\whline{1pt}%\cline{1-2}
	   \multirow{2}{*}{Methods} & \multicolumn{2}{c}{Supervision} & \multicolumn{2}{c}{mIoU (val)} & \multicolumn{1}{c}{mIoU (test)} \\ \cmidrule{2-3}\cmidrule(l){4-5}\cmidrule(l){6-6}
	    & weak & web & w/o CRF & w/ CRF & w/ CRF \\ \hline %\midrule[0.5pt]
        CCNN \cite{pathak2015constrained}, ICCV'15 & \cmark &
        	& 33.3\%& 35.3\%& - 	\\ %
        EM-Adapt \cite{papandreou2015weakly}, ICCV'15  & \cmark &
        	&  - 	& 38.2\%& 39.6\%\\ %
	    MIL \cite{pinheiro2015image}, CVPR'15 & \cmark &
	    	& 42.0\%& - 	& - 	\\ %
        SEC \cite{kolesnikov2016seed}, ECCV'16 & \cmark &
       	 	& 44.3\%& 50.7\%& 51.7\%\\ %
        AugFeed \cite{qi2016augmented}, ECCV'16 & \cmark &
        	& 50.4\%& 54.3\%& 55.5\%\\ %
        STC \cite{wei2016stc}, PAMI'16 & \cmark &
        	& - 	& 49.8\%& 51.2\%\\ %
        Roy et al. \cite{roycombining}, CVPR'17 & \cmark &
        	& - 	& 52.8\%& 53.7\%\\ %
        Oh et al. \cite{oh2017exploiting}, CVPR'17 & \cmark &
        	& 51.2\%& 55.7\%& 56.7\% \\ %
        AS-PSL \cite{wei2017object}, CVPR'17 & \cmark &
        	& - 	& 55.0\%& 55.7\% \\ %
        Hong et al. \cite{hong2017weakly}, CVPR'17 & \cmark &
        	& - 	& 58.1\% & 58.7\% \\
        WebS-i2 \cite{jin2017webly}, CVPR'17 & \cmark &
        	& - 	& 53.4\% & 55.3\% \\ %
        DCSP-VGG16 \cite{chaudhry2017discovering}, BMVC'17 & \cmark &
        	& 56.5\% & 58.6\% & 59.2\% \\
        Mining Pixels \cite{hou2016mining}, EMMCVPR'17 & \cmark &
        	& 56.9\%& 58.7\% & 59.6\% \\
        %DCSP-ResNet \cite{chaudhry2017discovering} & 59.5\% & 60.8\% & 60.3\% & 61.9\% \\
        $\text{WebSeg}^{\dagger}~\text{(Ours)}$ & \cmark &
        	& \textbf{60.5\%} & \textbf{63.1\%} & \textbf{63.3\%} \\  \midrule[0.4pt]
        WebSeg (Ours) & & \cmark
        	& 54.78\% & 57.10\% &  57.04\% \\ %\midrule[1pt]
        %Deeplab \cite{chen2014semantic} & & & 62.3\% & 67.6\% & 70.3\% \\ %
        \hline %\bottomrule[0.5pt]
        \end{tabular}
        %\vspace{5pt}
        \label{tab:comps}
        \vspace{-10pt}
\end{table}

\subsection{Discussion}

In Table~\ref{tab:comp_details}, we show the specific number of each category
on both the `val' and `test' sets .
As can be observed, with the supervision of accurate image-level labels, our approach
wins the other existing methods on most categories (the purple line).
However, there are also some unsatisfactory results for a few categories
(\eg `chair' and `table').
This small set of categories are normally with strange shapes and hence difficult to be
detected by current low-level cues extractors.
For instance, our method on the `table' category gets a IoU score of 24.1\%,
which is nearly 8\% worse than the AS-PSL method \cite{wei2017object}.
This is mainly due to the fact that AS-PSL mostly relies on the attention cues,
which performs better on detecting the location of semantic objects.

Regarding the results with web supervision, in spite of only several
low-level cues and extremely noisy web images, our approach can get comparable
results to the best weakly supervised methods.
More interestingly, our approach behaves the best on two categories on the `test' set
(`bicycle' and `train').
This implies that leveraging reliable low-level cues provides a promising way
to deal with learning semantic segmentation automatically from the Internet.

\myPara{Failure Case Analysis.}
Besides the successful samples, there are also many failure cases
that have been exhibited on the bottom part of Fig.~\ref{fig:visualComp}.
One of the main reasons leading to this should be the wrong predictions of the heuristic cues.
Although our online noise filtering mechanism, to some extent, is able to eliminate
the interference brought by noises, it still fails when processing some very complicated scenes.
For example, in Fig.~\ref{fig:illu_ignore}a, it is very hard to distinguish clearly
which regions belong to `bicycle' and which regions belong to `person.'
On the other hand, the simple images cannot incorporate all the scenes in the `test' set
and therefore results in many semantic regions undetected.

\newcommand{\xiaoliu}{\fontsize{6pt}{\baselineskip}\selectfont}
\begin{table*}[t]
    \centering
    \xiaoliu
    %\scriptsize
    \renewcommand{\arraystretch}{.8}
    \renewcommand{\tabcolsep}{0.45mm}
    \caption{Quantitative result comparisons with previous methods
    on each category. All the methods listed here are
    recent state-of-the-art methods. `${\dagger}$' means our methods with weak supervision.
    The best result in this column has been highlighted in \textbf{bold}.
    It can easily found that our approach based on weakly supervised setting works better
    than all the other methods on most categories.}
    \label{tab:comp_details}
    \vspace{5pt}
    \begin{tabular}{c|cccccccccccccccccccccc} \toprule[1.2pt]%\whline{1pt}%\cline{1-2}
        Methods &
        bkg &
        plane &
        bike &
        bird &
        boat &
        bottle &
        bus &
        car &
        cat &
        chair &
        cow &
        table &
        dog &
        horse &
        motor &
        person &
        plant &
        sheep &
        sofa &
        train &
        tv &
        mean \\ \midrule[.2mm]
        STC & 84.5 & 68.0 & 19.5 & 60.5 & 42.5 & 44.8 & 68.4 & 64.0 & 64.8 & 14.5 & 52.0 & 22.8 & 58.0 & 55.3 & 57.8 & 60.5 & 40.6 & 56.7 & 23.0 & 57.1 & 31.2 & 49.8 \\
        Roy et al. & 85.8 & 65.2 & 29.4 & 63.8 & 31.2 & 37.2 & 69.6 & 64.3 & 76.2 & 21.4 & 56.3 & 29.8 & 68.2 & 60.6 & 66.2 & 55.8 & 30.8 & 66.1 & \textbf{34.9} & 48.8 & 47.1 & 52.8 \\
        AS-PSL & 83.4 & 71.1 & 30.5 & 72.9 & 41.6 & 55.9 & 63.1 & 60.2 & 74.0 & 18.0 & 66.5 & \textbf{32.4} & 71.7 & 56.3 & 64.8 & 52.4 & 37.4 & 69.1 & 31.4 & 58.9 & 43.9 & 55.0 \\
        WebS-i2 & 84.3 & 65.3 & 27.4 & 65.4 & 53.9 & 46.3 & 70.1 & 69.8 & \textbf{79.4} & 13.8 & 61.1 & 17.4 & 73.8 & 58.1 & 57.8 & 56.2 & 35.7 & 66.5 & 22.0 & 50.1 & 46.2 & 53.4 \\
        Hong et al. & 87.0 & 69.3 & 32.2 & 70.2 & 31.2 & 58.4 & 73.6 & 68.5 & 76.5 & \textbf{26.8} & 63.8 & 29.1 & 73.5 & 69.5 & 66.5 & \textbf{70.4} & \textbf{46.8} & 72.1 & 27.3 & 57.4 & 50.2 & 58.1 \\
        MiningP & 88.6 & 76.1 & 30.0 & 71.1 & 62.0 & 58.0 & 79.2 & 70.5 & 72.6 & 18.1 & 65.8 & 22.3 & 68.4 & 63.5 & 64.7 & 60.0 & 41.5 & 71.7 & 29.0 & \textbf{72.5} & 47.3 & 58.7 \\
        \rowcolor{blue!8} \cellcolor{white!50} $\textbf{WebSeg}^{\dagger}$ & \textbf{89.6} & \textbf{85.6} & 32.3 & \textbf{76.1} & \textbf{68.3} & \textbf{68.7} & \textbf{84.2} & \textbf{74.9} & 78.2 & 18.6 & \textbf{75.0} & 24.1 & \textbf{75.2} & \textbf{69.9} & \textbf{67.0} & 63.9 & 40.8 & \textbf{77.9} & 32.8 & 69.7 & \textbf{53.3} & \textbf{63.1} \\

        \textbf{WebSeg} & 88.6 & 81.6 & \textbf{32.9} & 75.7 & 57.6 & 57.7 & 79.5 & 69.2 & 74.4 & 12.6 & 71.7 & 12.3 & 73.2 & 66.5 & 61.6 & 58.6 & 21.1 & 73.9 & 25.1 & 69.9 & 35.3 & 57.1 \\ \midrule[1.2pt]

        STC & 85.2 & 62.7 & 21.1 & 58.0 & 31.4 & 55.0 & 68.8 & 63.9 & 63.7 & 14.2 & 57.6 & 28.3 & 63.0 & 59.8 & 67.6 & 61.7 & 42.9 & 61.0 & 23.2 & 52.4 & 33.1 & 51.2 \\
        Roy et al. & 85.7 & 58.8 & 30.5 & 67.6 & 24.7 & 44.7 & 74.8 & 61.8 & 73.7 & 22.9 & 57.4 & 27.5 & 71.3 & 64.8 & 72.4 & 57.3 & 37.0 & 60.4 & \textbf{42.8} & 42.2 & 50.6 & 53.7 \\
        AS-PSL & - & - & - & - & - & - & - & - & - & - &- & - & - & - & - & - & - & - & - & - & - & 55.7 \\
        WebS-i2 & 85.8 & 66.1 & 30.0 & 64.1 & 47.9 & 58.6 & 70.7 & 68.5 & 75.2 & 11.3 & 62.6 & 19.0 & 75.6 & 67.2 & 72.8 & 61.4 & 44.7 & 71.5 & 23.1 & 42.3 & 43.6 & 55.3 \\
        Hong et al. & 87.2 & 63.9 & 32.8 & 72.4 & 26.7 & \textbf{64.0} & 72.1 & 70.5 & 77.8 & \textbf{23.9} & 63.6 & 32.1 & \textbf{77.2} & 75.3 & 76.2 & 71.5 & 45.0 & 68.8 & 35.5 & 46.2 & 49.3 & 58.7 \\
        MiningP & 88.9 & 72.7 & 31.0 & 76.3 & 47.7 & 59.2 & 74.3 & 73.2 & 71.7 & 19.9 & 67.1 & 34.0 & 70.3 & 66.6 & 74.4 & 60.2 & \textbf{48.1} & 73.1 & 27.8 & 66.9 & 47.9 & 59.6 \\
        \rowcolor{blue!8} \cellcolor{white!50} $\textbf{WebSeg}^{\dagger}$ & \textbf{89.8} & \textbf{78.6} & 32.4 & \textbf{82.9} & \textbf{52.9} & 61.5 & \textbf{79.8} & \textbf{77.0} & \textbf{76.8} & 18.8 & \textbf{75.7} & \textbf{34.1} & 75.3 & \textbf{75.9} & \textbf{77.1} & \textbf{65.7} & 46.1 & \textbf{78.9} & 32.3 & 65.3 & \textbf{52.8} & \textbf{63.3} \\

        \textbf{WebSeg} & 88.6 & 74.7 & \textbf{33.3} & 74.8 & 49.2 & 62.1 & 75.4 & 75.5 & 71.8 & 16.0 & 65.6 & 15.6 & 71.2 & 68.9 & 72.1 & 58.6 & 24.7 & 72.8 & 19.1 & \textbf{67.5} & 40.3 & 57.0 \\ \bottomrule[1.2pt]
        \end{tabular}
\end{table*}

\section{Conclusion and Discussion}

In this paper, we propose an interesting but challenging computer vision problem,
namely WebSeg, which aims at learning semantic segmentation from the free web images
following the learning manner of humans.
Regarding the extremely noisy web images and their imperfect proxy ground-truth annotations,
we design a novel online noise filtering mechanism, a new learning paradigm, to let CNNs
know how to discard undesired noisy regions.
Experiments show that our learning paradigm with only training on web images has already
obtained comparable results compared to previous state-of-the-art methods.
When leveraging more weak cues as in weakly supervised semantic segmentation,
we further improve the results by a large margin.
Moreover, we also perform a series of ablation experiments to show how each component
in our approach works.

Despite this, there is still a large room for improving the results of our task, which
is based on purely web supervision.
According to Table~\ref{tab:comp_details}, the mIoU scores of a few categories are still low.
Thereby, how to download good web images, improve the quality of heuristic cues, and design
useful noise-filtering mechanisms are interesting future directions.
This new topic, as mentioned above, covers a series of interesting but difficult techniques,
offering many valuable research directions which need to be further delved into.
In summary, automatically learning knowledge from the unrestricted Internet
resources substantially reduce the intervention of humans.
We hope such a interesting vision task could drive the rapid development of its relevant research topics as well.

{
\tiny
\bibliographystyle{splncs}
\bibliography{calc}
}
\end{document}